\documentclass[letterpaper]{article} 
\usepackage{aaai2026}  
\usepackage{times}  
\usepackage{helvet}  
\usepackage{courier}  
\usepackage[hyphens]{url}  
\usepackage{graphicx} 
\usepackage{natbib}  
\usepackage{caption} 
\usepackage{algorithm}
\usepackage{algorithmic}
\usepackage{graphicx} 
\usepackage{subcaption} 
\usepackage{multirow}
\usepackage{booktabs} 
\usepackage{amsmath}
\usepackage{amssymb}
\usepackage{diagbox}
\usepackage{arydshln}

\usepackage{newfloat}
\usepackage{listings}
\DeclareCaptionStyle{ruled}{labelfont=normalfont,labelsep=colon,strut=off} 
\floatstyle{ruled}
\newfloat{listing}{tb}{lst}{}
\floatname{listing}{Listing}
\title{Rectify Evaluation Preference: Improving LLMs' Critique on Math Reasoning \\ via Perplexity-aware Reinforcement Learning}

\author{
    Changyuan Tian\textsuperscript{\rm 1,\rm 2,\rm 3,}\equalcontrib, Zhicong Lu\textsuperscript{\rm 1,\rm 2,\rm 3}\equalcontrib, Shuang Qian\textsuperscript{\rm 4},
    Nayu Liu\textsuperscript{\rm 5}, Peiguang Li\textsuperscript{\rm 4,}\thanks{Corresponding authors.}, \\Li Jin\textsuperscript{\rm 1,\dag}, Leiyi Hu\textsuperscript{\rm 1,\rm 2,\rm 3}, Zhizhao Zeng\textsuperscript{\rm 4}, Sirui Wang\textsuperscript{\rm 4}, Ke Zeng\textsuperscript{\rm 4}, Zhi Guo\textsuperscript{\rm 1}
}

\affiliations {
    \textsuperscript{\rm 1}Aerospace Information Research Institute, Chinese Academy of Sciences\\
    \textsuperscript{\rm 2}Key Laboratory of Target Cognition and Application Technology (TCAT)\\
    \textsuperscript{\rm 3}School of Electronic, Electrical and Communication Engineering, University of Chinese Academy of Sciences\\
    \textsuperscript{\rm 4}Meituan\\ 
    \textsuperscript{\rm 5}School of Computer Science and Technology, Tiangong University\\
    tianchangyuan21@mails.ucas.edu.cn, jinlimails@gmail.com
}

\usepackage{bibentry}

\begin{document}

\maketitle

\begin{abstract}
To improve Multi-step Mathematical Reasoning (MsMR) of Large Language Models (LLMs), it is crucial to obtain scalable supervision from the corpus by automatically critiquing mistakes in the reasoning process of MsMR and rendering a final verdict of the problem-solution. Most existing methods rely on crafting high-quality supervised fine-tuning demonstrations for critiquing capability enhancement and pay little attention to delving into the underlying reason for the poor critiquing performance of LLMs. In this paper, we orthogonally quantify and investigate the potential reason — imbalanced evaluation preference, and conduct a statistical preference analysis. Motivated by the analysis of the reason, a novel perplexity-aware reinforcement learning algorithm is proposed to rectify the evaluation preference, elevating the critiquing capability. Specifically, to probe into LLMs' critiquing characteristics, a One-to-many Problem-Solution (OPS) benchmark is  meticulously constructed to quantify the behavior difference of LLMs when evaluating the problem solutions generated by itself and others. Then, to investigate the behavior difference in depth, we conduct a statistical preference analysis oriented on perplexity and find an intriguing phenomenon — "LLMs incline to judge solutions with lower perplexity as correct", which is dubbed as imbalanced evaluation preference. To rectify this preference, we regard perplexity as the baton in the algorithm of Group Relative Policy Optimization, supporting the LLMs to explore trajectories that judge lower perplexity as wrong and higher perplexity as correct. Extensive experimental results on our built OPS and existing available critic benchmarks demonstrate the validity of our method.
\end{abstract}


\section{Introduction}

Large Language Models (LLMs) have demonstrated exceptional capabilities in handling a wide range of tasks \cite{grattafiori2024llama, yang2025qwen3, lu-etal-2025-piper, liu-etal-2025-sara, lu-etal-2023-narrative, U-MERE}. However, their performance in Multi-step Mathematical Reasoning (MsMR) remains relatively limited. To improve the MsMR of LLMs, it is crucial to obtain scalable supervision from the corpus. Given the labor-intensive of directly annotating MsMR, researchers focus on employing instructed LLMs to automatically critiquing mistakes in the reasoning process of MsMR and rending a final verdict of the problem-solution.

However, the critiquing performance of instructed LLMs is far from meeting actual needs. To address this issue, existing works focus on crafting high-quality supervised fine-tuning demonstrations for critiquing capability enhancement. For instance, \citeauthor{wang-etal-2024-math} \shortcite{wang-etal-2024-math} adopt the monte carlo sampling to scalarly label the sample and achieve a critiquing model, which only judges the correctness of the entire problem-solution without critiquing process. To improve the interpretability of judgment, \citeauthor{DBLP:conf/acl/GaoCXWZLLLZXLC25} \shortcite{DBLP:conf/acl/GaoCXWZLLLZXLC25} leverages GPT-4 to finely label the intermediate step of sample, encouraging the critiquing model to identify the mistakes in the reasoning process. However, these data-driven methods pay little attention to delving into the underlying reason for the poor critiquing performance of instructed LLMs.

To bridge the gap of existing methods, our work is initiated with probing into LLMs' critiquing characteristics. Concretely, a One-to-many Problem Solution (OPS) benchmark is meticulously constructed, where each sample comprises one mathematical problem and many solutions generated by various families of LLMs (e.g., LLaMA, Qwen, and Mistral). Later, we quantify the behavior difference of LLMs when evaluating the problem-solution generated by itself and others. To investigate the behavior difference in depth, we conduct a statistical preference analysis oriented on perplexity and find an intriguing phenomenon — "LLMs incline to judge solutions with lower perplexity as correct", which is dubbed as the imbalanced evaluation preference. This biased preference hinders the model mining true cause-effect from supervised demonstrations during training (e.g., whether the final verdict of "correct" comes from lower perplexity or problem-solution itself.)

Motivated by the analysis of the potential reasons (i.e., the imbalanced evaluation preference) for the poor critiquing performance of LLMs, in this paper, we propose a novel perplexity-aware reinforcement learning algorithm (i.e., Group Relative Policy Optimization, GRPO) to rectify the evaluation preference. Specifically, we leverage perplexity as the baton to rescale the advantage distribution during Reinforcement Learning (RL), assigning greater weight to counter-preference trajectories (e.g., low perplexity but predicted wrong), thereby balancing LLMs' exploration. Additionally, we apply class-level loss aggregation by independently aggregating the losses for wrong-label and correct-label trajectories, ensuring that the rescaled comparative advantages within each class are faithfully reflected during optimization. These two designs jointly rectify evaluation preference and consequently improve the RL fine-tuning performance.

To validate the effectiveness of our method, we carry out extensive experiments on our built OPS and existing available critic benchmarks with mainstream LLMs. Compared to standard GRPO, our method alleviates the imbalanced evaluation preference and exhibits better critiquing capability. In summery, our main contributions include:

\begin{itemize}
    \item We meticulously construct a One-to-many Problem Solution (OPS) benchmark to quantify and  investigate the potential reason (i.e., imbalanced evaluation preference) for the poor critiquing performance of LLMs.

    \item Motivated by the analysis ("LLMs incline to judge solutions with lower perplexity as correct") of the potential reasons, we propose a novel perplexity-aware reinforcement learning algorithm to rectify the imbalanced evaluation preference, which supports the LLMs to explore counter-preference trajectories that judge lower perplexity as wrong and higher perplexity as correct.

    \item Extensive experiments on our built OPS benchmark demonstrate our method effectively alleviates the evaluation preference. The newly achieved state-of-the-art performance on existing available critic benchmarks further demonstrates the validity of our method.

\end{itemize}

\section{Analysis of Evaluation Preference}
In this section, we delve into the underlying reason for the poor critiquing performance of LLMs. To probe into the critiquing characteristics of LLMs, we meticulously construct a \textbf{O}ne-to-many \textbf{P}roblem-\textbf{S}olution (OPS) benchmark, to quantify the behavioral differences when LLMs evaluate problem solutions generated by themselves and by others. Furthermore, we conduct a statistical preference analysis based on perplexity and uncover an intriguing phenomenon: LLMs tend to judge solutions with lower perplexity as correct, a bias we refer to as \textit{imbalanced evaluation preference}.

\subsection{One-to-Many Problem-Solution Benchmark}
Unlike previous benchmarks that construct evaluation sets by randomly collecting mathematical reasoning solutions, our OPS benchmark specifically focuses on solutions generated by different families of LLMs that solve the same problem and arrive at the same final answer. This controlled design ensures that observed behavioral differences can be attributed to variations in reasoning processes.
\newline

\noindent \textbf{Mathematical Problems and Solutions Collection}. The mathematical problems used in OPS benchmark are sourced from the MATH \cite{DBLP:conf/nips/HendrycksBKABTS21} test dataset, which features a wide range of problem types and difficulty. To ensure diversity in the generated solutions, we employ three distinct LLMs—Qwen2-7B-Instruct, LLaMA3.1-8B-Instruct, and Mistral-7B-Instruct-v0.3—to independently solve each problem, using a chain-of-thought prompting strategy with a temperature setting of 0.7.  Solution correctness is labeled by comparing the final answer to the gold answer provided in the MATH dataset; a solution is labeled as \textit{correct} if the answers match, and \textit{wrong} otherwise. To further ensure data quality, we additionally leverage an advanced discriminative process reward model, Qwen2.5-Math-PRM-72B \cite{DBLP:conf/acl/ZhangZWZLYLZL25}, to filter the generated solutions. Solutions for which the correctness labels assigned by the process reward model and by answer matching are inconsistent are removed from the dataset.
\newline 

\noindent \textbf{Construction of Solution Triples}. 
To ensure that observed behavioral differences can be attributed to variations in reasoning processes, we construct solution triples. Each triple consists of independently generated solutions by Qwen, LLaMA, and Mistral for the same problem and yielding the same final answer. Formally, each triple is defined as:


\begin{equation}
T = \left[ x,\ \{y_1, y_2, y_3\},\ a,\ v \right]
\end{equation}
where $x$ denotes the mathematical problem, $y_i$ is the solution generated by the $i$-th model ($i \in \{1, 2, 3\}$ corresponding to Qwen, LLaMA, and Mistral), $a$ is the final answer extracted from $y_i$, $v \in \{\text{correct}, \text{wrong}\}$ indicates the correctness of $y_i$. To achieve a balanced evaluation, we ensure that the dataset contains an equal number of 'correct'-label and 'wrong'-label solutions (i.e., a 1:1 ratio).

Consequently, our OPS benchmark consists of three test subsets—LLaMA3.1-8B-Instruct, Mistral-7B-Instruct-v0.3, and Qwen2-7B-Instruct—each containing 630 samples with the same problems and final answers, resulting in a total of 1,890 test samples.
\newline

\noindent \textbf{Metrics}. To measure evaluation preference, we introduce the \textit{Balance Indicator} (BI), defined as the difference between the false positive rate (FPR) and the false negative rate (FNR):

\begin{equation}
\text{BI} = \text{FPR} - \text{FNR}     
\end{equation}
here, FPR is the proportion of actually incorrect solutions erroneously judged as correct, while FNR is the proportion of actually correct solutions erroneously judged as wrong. 
A BI close to zero indicates balanced evaluation preference; positive BI suggests overestimation of incorrect solutions, and negative BI suggests underestimation of correct ones. The BI ranges from $-1$ (maximal underestimation) to $1$ (maximal overestimation), with $0$ indicating perfect balance. In addition, we also use basic accuracy as another metric.

\subsection{Imbalanced Evaluation Preference}
\begin{table}[htbp]
\centering
\small
\setlength{\tabcolsep}{0.6mm}
\begin{tabular}{lcccccc}
\toprule
\multirow{2}{*}{Model} 
& \multicolumn{2}{c}{L-subset} 
& \multicolumn{2}{c}{M-subset} 
& \multicolumn{2}{c}{Q-subset} \\
\cmidrule(lr){2-3} \cmidrule(lr){4-5} \cmidrule(lr){6-7}
& Acc & BI & Acc & BI & Acc & BI \\
\midrule

Qwen2-7B-Instruct 
& 65.56 & 34.60 
& 67.94 & 2.54 
& 60.63 & 66.03 \\

LLaMA3.1-8B-Instruct 
& 64.13 & 17.78 
& 62.86 & -30.48 
& 65.24 & 16.83 \\

Mistral-7B-Instruct-v0.3 
& 52.70 & 81.27 
& 51.90 & 86.03 
& 52.54 & 86.67 \\
\bottomrule
\end{tabular}
\caption{Performance of each model on self and cross-evaluation per subset. The L-subset, M-subset, and Q-subset columns correspond to evaluation results on the LLaMA3.1-8B-Instruct, Mistral-7B-Instruct-v0.3, and Qwen2-7B-Instruct test subsets, respectively.}
\label{tab:self-overestimation}
\end{table}

\noindent Based on OPS benchmark, we evaluate Qwen2-7B-Instruct, LLaMA3.1-8B-Instruct, and Mistral-7B-Instruct-v0.3, with the results presented in Table \ref{tab:self-overestimation}. 
Firstly, a direct observation is that despite the three subsets sharing the same problems and final answers, the three LLMs exhibit markedly different evaluation performance across the subsets. This suggests that evaluation behavior may be affected by model-specific reasoning, leading to inconsistent assessments across solutions generated by different models. Furthermore, we observe a clear imbalanced evaluation preference in LLMs: each model tends to achieve a higher and more positive BI when evaluating its own solutions, while BI drops substantially when evaluating solutions from other models (e.g., Qwen: self $66.03$ vs. cross $2.54$, LLaMA: self $17.78$ vs. cross $-30.48$, Mistral: self $86.03$ vs. cross $81.27$). This imbalanced preference impairs model evaluation performance.

\subsection{Statistical Preference Analysis}
Motivated by the observed evaluation preference, we further conduct a fine-grained statistical analysis by exploring the correlation between perplexity and BI. Perplexity reflects how closely a solution aligns with the critic model’s own generation style, with lower values indicating smaller textual divergence. Specifically, for each critic model, we calculate the perplexity of each problem-solution pair in OPS benchmark, partition the data into decile bins based on perplexity, and perform linear regression analysis to investigate the relationship between perplexity and BI.

\begin{figure}[htbp]
    \centering
    \includegraphics[width=0.47\textwidth]{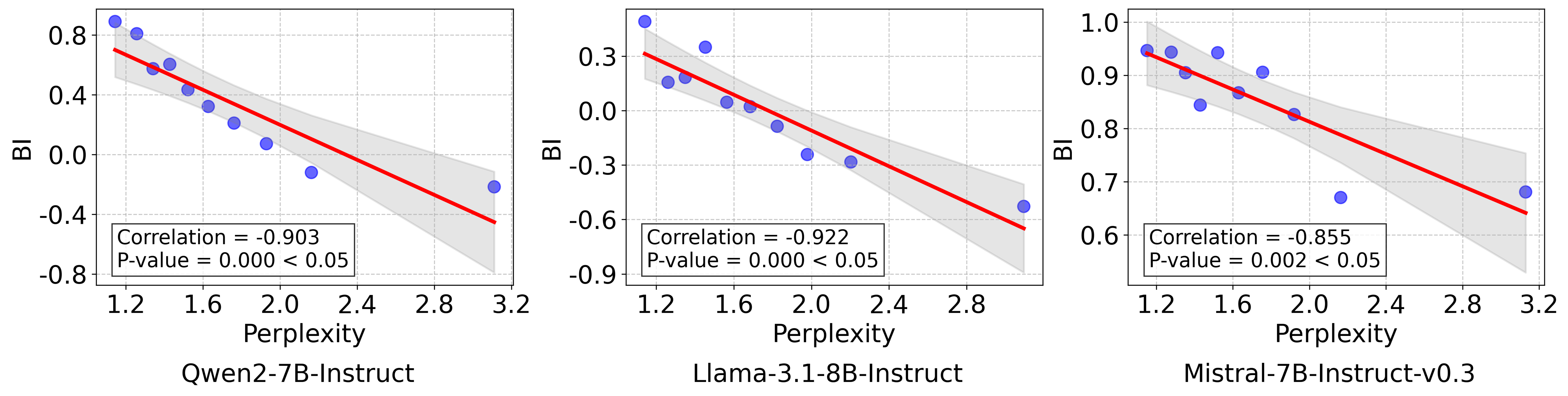}
    \caption{Visualization of the negative correlation between perplexity and BI for each critic model. Each point represents a decile bin, with the red regression line and gray-shaded area denoting the fitted linear trend and the $95\%$ confidence interval, respectively. Higher perplexity values are associated with lower BI.}
    \label{fig:f1}
\end{figure}

We observe a clear linear negative correlation between perplexity and BI (see Figure~\ref{fig:f1}): as perplexity increases—indicating that a solution deviates more from the critic model’s generation style—the critic model is more likely to judge it as wrong, resulting in BI values shifting toward $-1$. This pattern holds consistently across all three models, each exhibiting a significant negative correlation coefficient with $p$-values less than 0.05. The strong correlation between evaluation preference and perplexity suggests a promising entry point for us to manipulate and mitigate such bias.

\begin{figure*}
    \centering
    \includegraphics[width=1\linewidth]{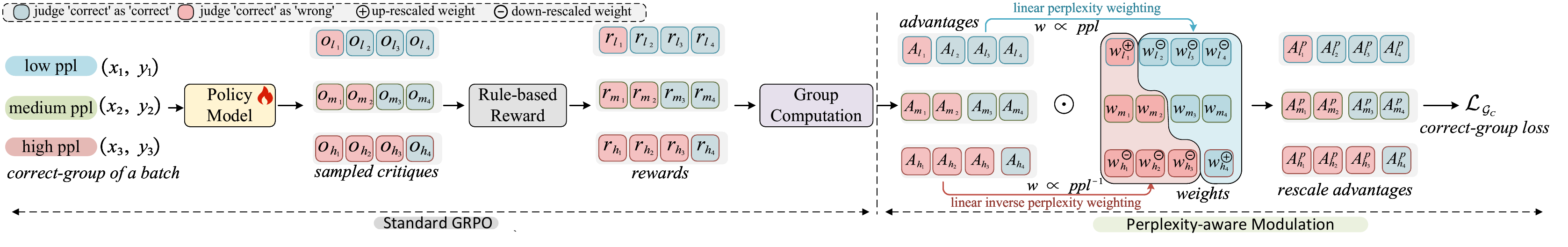}
    \caption{Illustration of our proposed perplexity-aware GRPO. Here, we take the correct-group within a batch (i.e., the sample set where the ground truth is correct) as an example; the wrong group follows the same process.}
    \label{fig:main}
\end{figure*}
\section{Method}
\subsection{Problem Setup}

Given a mathematical problem $x$ and its proposed solution $y$, our model is designed to provide a comprehensive critique of $y$. Specifically, the model outputs a triplet $(f, v, \hat{e})$, where $f$ denotes the step-by-step critique on each reasoning step in $y$, $v \in \{\text{correct},\, \text{wrong}\}$ indicates the overall correctness of the solution, and $\hat{e}$ records the content of the first erroneous step, if any. If all steps in $y$ are correct, then $v = \text{correct}$ and $\hat{e} = \varnothing$; otherwise, $v = \text{wrong}$ and $\hat{e}$ contains the first incorrect step identified by the model. Formally, this evaluation process can be expressed as:

\begin{equation}
(f, v, \hat{e}) = LLM(x, y)    
\end{equation}
where $LLM(\cdot)$ denotes the evaluation model. This formulation enables fine-grained analysis of the solution process, supporting both detailed diagnostic feedback and precise error localization.

\subsection{Training Data Curation}

The mathematical problems used in our study are sourced from the MATH \cite{DBLP:conf/nips/HendrycksBKABTS21} train dataset, which covers diverse problem types and difficulty levels. Consistent with the OPS benchmark, we employ the same three LLMs—Qwen2-7B-Instruct, LLaMA3.1-8B-Instruct, and Mistral-7B-Instruct-v0.3—for solution sampling, ensuring a wide range of solution perplexities.

A key requirement for our training data curation is the accurate identification of the first erroneous step within each solution. Following the ProcessBenchmark methodology \cite{zheng2025processbenchidentifyingprocesserrors}, we annotate each solution as follows: first, Qwen2.5-72B-Instruct \cite{qwen2025qwen25technicalreport} is used to segment the solution into individual reasoning steps. Each step is then evaluated by the state-of-the-art process reward model, Qwen2.5-Math-PRM-72B \cite{DBLP:conf/acl/ZhangZWZLYLZL25}. Solutions that exhibit inconsistencies between stepwise and overall correctness are filtered out to maintain annotation quality. The first step with a reward score below 0.8 is labeled as the initial error.

Formally, each training example is represented as $D = \left\{x, y, a, v, e^*\right\}$, where $x$ is the mathematical problem, $y$ is the solution generated by the model (Qwen, LLaMA, or Mistral), $a$ is the gold answer, $v \in \{\text{correct}, \text{wrong}\}$ denotes the correctness of $y$, and $e^*$ is the first erroneous step in $y$.

The final training set comprises 5,760 samples, with equal numbers of solutions sampled from each of the three models. For each model, correct and incorrect solutions are balanced, ensuring a well-balanced training dataset.

\subsection{Perplexity-aware Reinforcement Learning}
To rectify evaluation preference, our approach leverages perplexity to modulate the advantage distribution during reinforcement learning, thereby achieving more balanced evaluation performance. Without loss of generality, we adopt the well-established GRPO \cite{DBLP:journals/corr/abs-2402-03300} method as our base, which does not require step-by-step supervision signals. An illustration of our proposed perplexity-aware GRPO is shown in Figure \ref{fig:main}.
\newline

\noindent \textbf{Standard GRPO.}  GRPO first samples a group of candidate critiques $\{o_i\}_{i=1}^G$ for each question-answer pair, assigns reward scores $r_{i}$ to these critiques, and then estimates their advantages $A_{i}$ by normalizing the group-level rewards. Its objective function for each critique is defined as: 

\begin{equation}
\begin{aligned}
\mathcal{J}(\theta) = \frac{1}{|o_i|} \sum_{t=1}^{|o_i|} \Bigg[ \min \Big( r_{i,t}(\theta) A_{i,t},\; \text{clip}\big(r_{i,t}(\theta),\, \\
1-\varepsilon, 
1+\varepsilon\big)\, A_{i,t} \Big) - \beta\, D_{\text{KL}}\big(\pi_\theta \| \pi_{\text{ref}}\big) \Bigg]
\end{aligned}    
\end{equation}
here, $r_{i,t}(\theta)$ denotes the ratio of the current policy’s probability to the old policy’s probability for the $t$-th token in the $i$-th critique. The advantage $A_{i}$ is computed by normalizing the $i$-th critique's reward $r_i$ against the mean and standard deviation of rewards across all $G$ critiques in the group:  
$A_{i} = \frac{r_i - \text{mean}(\{r_i\}_{i=1}^G)}{\text{std}(\{r_i\}_{i=1}^G)}$. 
\newline

\noindent \textbf{Perplexity-aware Modulation.} The prior analysis reveals that LLMs tend to evaluate solutions with lower perplexity as correct, and those with higher perplexity as incorrect. This imbalanced evaluation preference narrows the LLMs' potential trajectory exploration space: they rarely explore trajectories that identify high-perplexity solutions as correct or low-perplexity ones as wrong. Such insufficient and  imbalanced exploration ultimately leads to suboptimal performance. 

To promote more balanced exploration, we introduce a perplexity-aware modulation mechanism. The core idea is to upweight high-perplexity-but-predicted-correct and low-perplexity-but-predicted-wrong cases, thereby encouraging the model to explore counter-preference behaviors in both directions. First, each training batch is split into two groups based on the ground-truth labels: the correct-group $\mathcal{G}_{\text{C}}$, containing cases labeled as correct, and the wrong-group $\mathcal{G}_{\text{W}}$, containing cases labeled as wrong. Within each group (e.g., $\mathcal{G}_{\text{C}}$), we further partition the data into subgroups according to the model’s predictions: $\mathcal{G}_{\text{C}}^{\text{c}}$ for cases predicted as correct, and $\mathcal{G}_{\text{C}}^{\text{w}}$ for cases predicted as wrong. For subgroups $\mathcal{G}_{\text{C}}^{\text{c}}$ and $\mathcal{G}_{\text{W}}^{\text{c}}$, we employ a linear, perplexity-aware advantage modulation, assigning greater weights (i.e., $>1$) to samples with higher perplexity. Conversely, for subgroups $\mathcal{G}_{\text{C}}^{\text{w}}$ and $\mathcal{G}_{\text{W}}^{\text{w}}$, we apply an inverse linear modulation, whereby lower perplexity corresponds to greater weights.

Formally, take $\mathcal{G}_{\text{C}}$ as an example. Let $ppl$  denote the perplexity of a problem-solution pair within this group. The weight assigned to each corresponding critique is given by
\begin{equation}
w_i =
\begin{cases}
ppl_i / \text{mean}(\{ppl_k \mid k \in \mathcal{G}_{\text{C}}^{\text{c}}\}), & i \in \mathcal{G}_{\text{C}}^{\text{c}} \\[8pt]
\text{mean}(\{ppl_k \mid k \in \mathcal{G}_{\text{C}}^{\text{w}}\}) / ppl_i, & i \in \mathcal{G}_{\text{C}}^{\text{w}}
\end{cases}    
\end{equation}
where $\mathcal{G}_{\text{C}}^{\text{c}}$ and $\mathcal{G}_{\text{C}}^{\text{w}}$ denote the subgroups of $\mathcal{G}_{\text{C}}$ predicted as correct and wrong, respectively.

These modulation weights $w_i$ are then used to rescale the advantages for each critique prior to policy optimization:

\begin{equation}
A_{i}^{\text{p}} = w_i \cdot A_{i}.
\end{equation}
this perplexity-aware modulation encourages the policy to cover less-explored trajectory spaces, mitigating the LLM's bias and thereby enhancing RL fine-tuning performance.
\newline

\noindent \textbf{Class-Level Loss Aggregation}. In vanilla GRPO, the loss for each sample in a batch is aggregated by taking the overall mean, i.e., 
$\mathcal{L} = \frac{1}{B} \sum_{i=1}^{B} \mathcal{J}_i.$
In our method, however, the advantage distribution within each training batch is modulated separately within groups defined by the ground truth: the correct group ($\mathcal{G}_C$) and the wrong group ($\mathcal{G}_W$). To ensure that the perplexity-modulated comparative advantages are fully reflected in the loss, we apply class-level loss aggregation. Specifically, we first compute the mean of nonzero loss terms within each group, and then average these group-wise means to obtain the final loss. Formally, the class-level loss is defined as:

\begin{equation}
\mathcal{L}^\prime = \frac{1}{2} \left( \frac{1}{|\mathcal{G}_C'|} \sum_{i \in \mathcal{G}_C'} \mathcal{J}_i + \frac{1}{|\mathcal{G}_W'|} \sum_{i \in \mathcal{G}_W'} \mathcal{J}_i \right),    
\end{equation}
where $\mathcal{G}_C' = \{i \in \mathcal{G}_C \mid \mathcal{J}_i \neq 0\}$ and $\mathcal{G}_W' = \{i \in \mathcal{G}_W \mid \mathcal{J}_i \neq 0\}$ denote the sets of samples with nonzero loss in the correct and wrong groups, respectively.
\newline

\noindent \textbf{Reward Design.} We design a rule-based reward function to guide the model toward generating accurate verdicts and identifying the first erroneous step. The overall reward comprises two components: the format reward $r_{\text{f}}$ and the answer reward $r_{\text{a}}$. The format reward assigns $0.1$ if the response strictly follows the predefined format, and $0$ otherwise.

The answer reward $r_{\text{a}}$ reflects the hierarchical structure of the mathematical reasoning evaluation task. For actually incorrect problem-solution pairs ($v^* = \text{'wrong'}$), partial reward is given for correctly judging the solution as wrong, with additional reward for accurately localizing the first erroneous step. For actually correct pairs ($v^* = \text{'correct'}$), the full reward is granted only for correctly judging the solution as correct.
Specifically, the answer reward $r_{\text{a}}$ is defined as:

\begin{equation}
r_{\text{a}} =
\begin{cases}
0.8, & v^* = \text{'correct'} \text{ and } v = \text{'correct'} \\
0.6 + \mathrm{F1}(\hat{e}, e^*), & v^* = \text{'wrong'} \text{ and } v = \text{'wrong'} \\
0, & \text{otherwise}
\end{cases}    
\end{equation}
where $\mathrm{F1}(\hat{e}, e^*)$ denotes the F1 score between the predicted and ground truth first erroneous steps. To enhance reward stability and eliminate the influence of minor fluctuations, we round the F1 score to one decimal place, thereby suppressing variations smaller than $0.1$. Note that directly employing a simple binary (0-1) reward incentivizes the model to overpredict 'correct' due to the relative ease of obtaining full reward for correct solutions, while failing to promote precise identification of the first erroneous step. To avoid such shortcut behavior, the reward function is designed with a graded scheme (with base rewards such as 0.8 and 0.6 empirically chosen within a typical range) that aligns reward magnitude with task difficulty. The final reward $r$ is computed as $r_{\text{f}} + r_{\text{a}}$. 

\section{Experiments}
\subsection{Datasets}
In our experiments, we employ two benchmarks: our proposed OPS benchmark and the public ProcessBenchmark \cite{zheng2025processbenchidentifyingprocesserrors}. The OPS benchmark comprises solutions to MATH test data problems generated by Qwen2-7B-Instruct \cite{team2024qwen2}, LLaMA3.1-8B-Instruct \cite{grattafiori2024llama}, and Mistral-v0.3-Instruct \cite{jiang2023mistral7b}. All subsets share the same problems and final answers, but differ in their solution processes. ProcessBench assesses a model’s ability to localize errors in mathematical reasoning and consists of four subsets—GSM8K, MATH, OlympiadBench, and Omni-MATH—with solution processes sampled from various models.

\subsection{Baselines}  

We employ three mainstream open-source aligned models: Qwen2-7B-Instruct, LLaMA3.1-8B-Instruct, and Mistral-7B-Instruct-v0.3, to validate the effectiveness of our proposed method. Furthermore, we incorporate SFT, standard GRPO \cite{DBLP:journals/corr/abs-2402-03300}, and DrGRPO \cite{liu2025understanding} variants as baseline models to facilitate comparison with our perplexity-aware GRPO approach. For SFT, step-by-step supervision is provided by annotations from GPT-4o.

\subsection{Experimental Settings}

The implementations of the SFT, GRPO, and DrGRPO baselines are based on the publicly available VeRL \cite{DBLP:conf/eurosys/ShengZYWZZPL025} open-source project. Likewise, our proposed perplexity-aware GRPO method is also developed on top of VeRL. All experiments are conducted using eight NVIDIA A100-80G GPUs with bfloat16 precision. For the training of GRPO, DrGRPO, and Perplexity-aware GRPO, we set the learning rate to \(1e-6\) and use a train batch size of 128. The total number of training steps is set to 400. During the rollout phase, 5 samples are generated for each prompt with a temperature of 1.0. Rollout generation is support by vLLM \cite{kwon2023efficientmemorymanagementlarge}. We save checkpoints every 50 steps during training and select the checkpoint with the best average performance across the two benchmarks. For SFT, we train for 3 epochs with a learning rate of $2e-6$ and select the checkpoint with the best average performance. All evaluations use greedy decoding (temperature=0).

\begin{table*}[ht]
\centering
\small
\setlength{\tabcolsep}{1mm}
\begin{tabular}{lcccccccccccccc}
\toprule
\multirow{2}{*}{Model} 
& \multicolumn{4}{c}{LLaMA3.1-8B-Instruct-Subset} 
& \multicolumn{4}{c}{Mistral-7B-Instruct-v0.3-Subset} 
& \multicolumn{4}{c}{Qwen2-7B-Instruct-Subset} 
& \multicolumn{2}{c}{\textit{Average}} \\
\cmidrule(lr){2-5} \cmidrule(lr){6-9} \cmidrule(lr){10-13} \cmidrule(lr){14-15}
& FPR & FNR & BI & Acc
& FPR & FNR & BI & Acc
& FPR & FNR & BI & Acc
& \textit{$\vert$BI$\vert$~$\downarrow$} & \textit{Acc~$\uparrow$} \\

\midrule

Qwen2-7B-Instruct 
& 51.75 & 17.14 & 34.60 & 65.56
& 33.33 & 30.79 & 2.54 & 67.94
& 72.38 & 6.35 & 66.03 & 60.63
& \textit{34.39} & \textit{64.71} \\

\cdashline{1-15}

+ SFT 
& 31.11 & 26.98 & 4.13 & 70.95
& 16.83 & 49.84 & -33.02 & 66.67
& 38.73 & 22.54 & 16.19 & 69.37
& \underline{\textit{17.78}} & \textit{69.00} \\
+ Vanilla GRPO 
& 7.94 & 39.05 & -31.11 & 76.51
& 3.17 & 54.92 & -51.75 & 70.95
& 11.75 & 34.60 & -22.86 & 76.83
& \textit{35.24} & \textit{74.76} \\

+ DrGRPO 
& 9.52 & 33.33 & -23.81 & 78.57
& 3.81 & 48.25 & -44.44 & 73.97
& 10.79 & 31.43 & -20.63 & 78.89
& \textit{29.63} & \underline{\textit{77.14}} \\

+ Perplexity-aware GRPO 
& 14.29 & 27.62 & -13.33 & 79.05  
& 5.71 & 33.97 & -28.25 & 80.16  
& 14.92 & 24.76 & -9.84 & 80.16
& \textbf{\textit{17.14}} & \textbf{\textit{79.79}} \\

\midrule

LLaMA3.1-8B-Instruct 
& 44.76 & 26.98 & 17.78 & 64.13 
& 21.90 & 52.38 & -30.48 & 62.86 
& 43.17 & 26.35 & 16.83 & 65.24 
& \textit{21.70} & \textit{64.07} \\

\cdashline{1-15}

+ SFT 
& 31.11 & 25.71 & 5.40 & 71.59 
& 9.52 & 49.21 & -39.68 & 70.63 
& 31.75 & 23.81 & 7.94 & 72.22 
& \underline{\textit{17.67}} & \textit{71.48} \\

+ Vanilla GRPO 
& 6.98 & 39.68 & -32.70 & 76.67 
& 5.08 & 48.25 & -43.17 & 73.33 
& 6.35 & 40.32 & -33.97 & 76.67 
& \textit{36.61} & \textit{75.56} \\

+ DrGRPO  
& 6.98 & 38.10 & -31.11 & 77.46  
& 6.98 & 44.44 & -37.46 & 74.29  
& 7.94 & 35.24 & -27.30 & 78.41
& \textit{31.96} & \underline{\textit{76.72}} \\

+ Perplexity-aware GRPO 
& 18.10 & 26.67 & -8.57 & 77.62  
& 8.57 & 37.46 & -28.89 & 76.98  
& 14.29 & 27.30 & -13.02 & 79.21
& \textbf{\textit{16.83}} & \textbf{\textit{77.94}} \\

\midrule

Mistral-7B-Instruct-v0.3 
& 87.94 & 6.67 & 81.27 & 52.70
& 91.11 & 5.08 & 86.03 & 51.90
& 90.79 & 4.13 & 86.67 & 52.54
& \textit{84.65} & \textit{52.38} \\

\cdashline{1-15}

+ SFT 
& 28.25 & 31.43 & -3.17 & 70.16  
& 20.32 & 43.49 & -23.17 & 68.10  
& 40.63 & 26.03 & 14.60 & 66.67
& \textbf{\textit{13.65}} & \textit{68.31} \\

+ Vanilla GRPO 
& 14.29 & 44.44 & -30.16 & 70.63  
& 13.65 & 53.65 & -40.00 & 66.35  
& 17.78 & 45.71 & -27.94 & 68.25
& \textit{32.70} & \underline{\textit{68.41}} \\

+ DrGRPO 
& 13.02 & 49.21 & -36.19 & 68.89
& 11.43 & 61.59 & -50.16 & 63.49
& 13.97 & 49.52 & -35.56 & 68.25
& \textit{40.64} & \textit{66.88} \\

+ Perplexity-aware GRPO 
& 16.51 & 44.13 & -27.62 & 69.68  
& 17.78 & 47.94 & -30.16 & 67.14  
& 22.22 & 38.41 & -16.19 & 69.68
& \underline{\textit{24.66}} & \textbf{\textit{68.83}} \\

\bottomrule
\end{tabular}
\caption{Performance comparison on the OPS benchmark. FPR: false positive rate; FNR: false negative rate; BI: signed balance indicator; $\vert$BI$\vert$: absolute BI. Best results are in bold; second-best are underlined.}
\label{tab:one-to-many}
\end{table*}

\subsection{Experimental Results}

The experimental results on our proposed OPS benchmark are presented in Table \ref{tab:one-to-many}. Our perplexity-aware GRPO method outperforms the baseline approaches in terms of both accuracy and balance indicator in most cases across all three base models. These results demonstrate the effectiveness of our method in rectifying evaluation preference and thereby improving RL fine-tuning performance, as well as its generalizability across different model series.

Further, we observe that: (1) SFT demonstrates robust and consistent improvements (e.g., Qwen: 69.00, LLaMA: 71.48, Mistral: 68.31) across models with varying initial performance, whereas RL methods such as GRPO are more sensitive to the base model’s starting accuracy and do not offer clear advantages over SFT when the initial performance is low (e.g., Mistral-7B-Instruct-v0.3: starting accuracy 52.38, close to random guess; after RL, 68.41, similar to SFT’s 68.31). This suggests that the initial behavior of the base model, such as its evaluation preference, may affect the effectiveness of subsequent RL fine-tuning. (2) For strong base models such as Qwen2 and LLaMA3.1, the GRPO and DrGRPO methods achieve remarkable performance gains over SFT by leveraging the exploration-exploitation capabilities of RL; however, they face severe imbalance issue—larger $\vert$BI$\vert$ and noticeable performance differences across subsets—resulting in suboptimal performance. (3) In contrast, our perplexity-aware GRPO delivers more balanced and superior evaluation results, with smaller $\vert$BI$\vert$ and reduced performance differences across subsets. This highlights the effectiveness of using perplexity to calibrate imbalanced advantage distributions.

The experimental results on the public ProcessBench dataset are summarized in Table \ref{tab:ProcessBench}, where the F1 score is reported based on the accuracies of label-wrong and label-correct samples. Considering ProcessBench’s requirement to precisely localize the first erroneous step, our proposed perplexity-aware GRPO achieves superior performance in most cases (11 out of 15) across three base models. These results further demonstrate that our method substantially enhances critiquing capability, yielding consistent improvements across four datasets of varying difficulty. Notably, the SFT method lags significantly behind other RL methods, indicating poor error step localization capability, which may be related to the challenge of obtaining large-scale, accurately labeled step-by-step supervision data. In contrast, vanilla GRPO and DrGRPO tend to overfit the training data stem from MATH, as reflected by their high performance on MATH (e.g., LLaMA3.1-8B-Instruct + GRPO achieves 43.38), but they struggle to generalize to other datasets, resulting in suboptimal overall performance. By rectifying imbalanced preference, our perplexity-aware GRPO demonstrates more balanced cross-dataset generalization and achieves the best overall performance.

\subsection{Ablation Study}
To further evaluate the effectiveness of our method, we conduct ablation studies on the OPS benchmark using Qwen2-7B-Instruct as the base model. As shown in Table~\ref{tab:ab}, we consider the following three ablation variants: (1) \textit{-w/o class-level aggregation}, which disables class-level aggregation and reverts to vanilla aggregation; (2) \textit{-w/o perplexity modulation}, which removes perplexity-aware modulation; and (3) \textit{-w/o both}, which disables both components and is equivalent to the vanilla GRPO.
We make the following observations: (1) removing class-level aggregation leads to an increase in $\vert$BI$\vert$ and a slight decrease in accuracy, indicating that class-level aggregation plays a positive role in maintaining optimization balance. (2) in contrast, removing perplexity modulation results in a significant drop in accuracy, even though it noticeably reduces $\vert$BI$\vert$. This suggests that simply minimizing $\vert$BI$\vert$ through class-level aggregation does not directly result in substantial performance improvement, as it fails to tackle the core issue of imbalanced exploitation during RL. Fundamentally, leveraging perplexity to balance the model’s exploration is crucial for improving RL performance. (3) disabling both components leads to the worst performance, underscoring the necessity of each core design.

\begin{table}[htbp]
\centering
\small
\setlength{\tabcolsep}{1mm} 
\begin{tabular}{lccccc}
\toprule
Model & {GM} & {MT} & {OB} & {OM} & {AVG} \\
\midrule
Qwen2-7B-Instruct & 24.97 & 31.47 & 27.36 & 22.41 & 26.55 \\
\cdashline{1-6}
+ SFT & 26.00 & 26.64 & 25.17 & 20.74 & 24.64 \\
+ Vanilla GRPO & \underline{45.24} & 42.12 & 32.93 & 31.52 & 37.95 \\
+ DrGRPO & 42.03 & \underline{43.45} & \underline{36.79} & \textbf{38.69} & \underline{40.24} \\
+ Perplexity-aware GRPO & \textbf{48.05} & \textbf{43.91} & \textbf{40.31} & \underline{36.68} & \textbf{42.24} \\

\midrule
LLaMA3.1-8B-Instruct & 24.68 & 12.88 & 11.10 & 12.45 & 15.28 \\
\cdashline{1-6}
+ SFT & 31.00 & 29.12 & 27.27 & 23.46 & 27.71 \\
+ Vanilla GRPO & \underline{57.88} & \textbf{43.38} & 25.33 & 29.06 & 38.91 \\
+ DrGRPO & 55.60 & 40.64 & \underline{31.80} & \underline{33.67} & \underline{40.43} \\
+ Perplexity-aware GRPO & \textbf{59.82} & \underline{41.32} & \textbf{33.96} & \textbf{35.15} & \textbf{42.56} \\

\midrule
Mistral-7B-Instruct-v0.3 & 15.65 & 18.18 & 10.34 & 10.64 & 13.70\\
\cdashline{1-6}
+ SFT & \underline{16.85} & 23.47 & 22.27 & 22.16 & 21.19 \\
+ Vanilla GRPO & 13.40 & 34.77 & \underline{31.33} & 28.85 & 27.09 \\
+ DrGRPO & 12.73 & \textbf{37.64} & \textbf{32.08} & \underline{29.88} & \underline{28.08} \\
+ Perplexity-aware GRPO & \textbf{19.09} & \underline{35.57} & 28.48 & \textbf{30.66} & \textbf{28.45} \\
\bottomrule
\end{tabular}
\caption{Evaluation results on ProcessBench. GM, MT, OB, and OM correspond to the abbreviations of GSM8K, MATH, OlympiadBench, and Omni-MATH, respectively. 
}
\label{tab:ProcessBench}
\end{table}

\begin{figure}[htbp]
    \centering
    \includegraphics[width=0.46\textwidth]{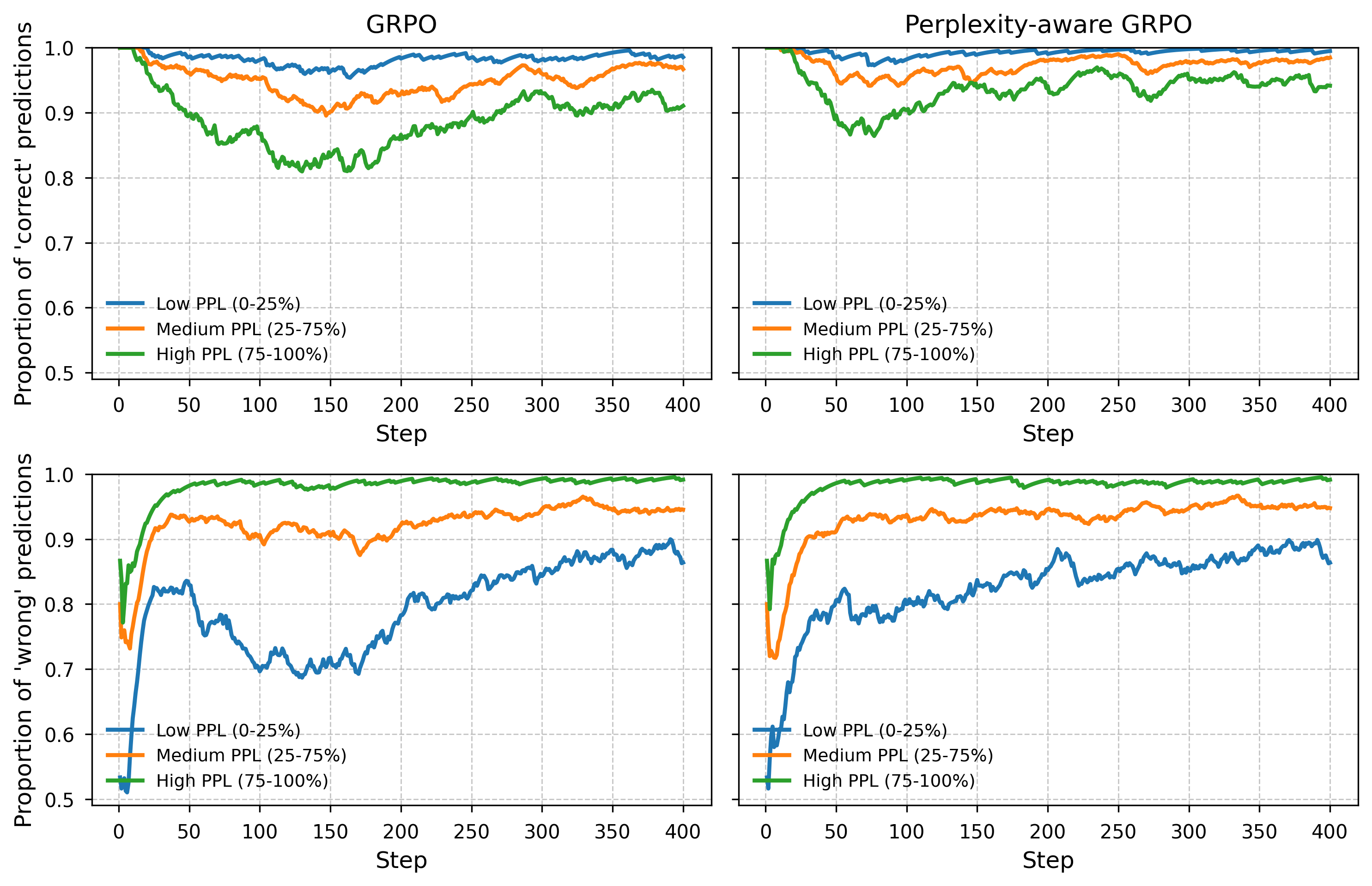}
    \caption{Exploration comparison between vanilla and perplexity-aware GRPO. Closer curves across perplexity bins imply more balanced exploration.}
    \label{fig:dynamic}
\end{figure}

\begin{table*}[ht]
\centering
\small
\setlength{\tabcolsep}{1mm} 
\begin{tabular}{lcccccccccccccc}
\toprule
\multirow{2}{*}{Model} 
& \multicolumn{4}{c}{LLaMA3.1-8B-Instruct-Subset} 
& \multicolumn{4}{c}{Mistral-7B-Instruct-v0.3-Subset} 
& \multicolumn{4}{c}{Qwen2-7B-Instruct-Subset} 
& \multicolumn{2}{c}{\textit{Average}} \\
\cmidrule(lr){2-5} \cmidrule(lr){6-9} \cmidrule(lr){10-13} \cmidrule(lr){14-15}
& FPR & FNR & BI & Acc
& FPR & FNR & BI & Acc
& FPR & FNR & BI & Acc
& \textit{$\vert$BI$\vert$~$\downarrow$} & \textit{Acc~$\uparrow$} \\

\midrule

Perplexity-aware GRPO & 14.29 & 27.62 & -13.33 & 79.05 & 5.71 & 33.97 & -28.25 & 80.16 & 14.92 & 24.76 & -9.84 & 80.16 & \textit{\underline{17.14}} & \textit{\textbf{79.79}} \\

\cdashline{1-15}

-w/o class-level aggregation & 7.62 & 27.94 & -20.32 & 82.22 & 4.76 & 42.22 & -37.46 & 76.51 & 15.87 & 30.48 & -14.60 & 76.83 & \textit{24.13} & \textit{\underline{78.52}} \\

-w/o perplexity modulation & 24.13 & 20.00 & 4.13 & 77.94 & 14.29 & 34.29 & -20.00 & 75.71 & 31.75 & 20.95 & 10.79 & 73.65 & \textit{\textbf{11.64}} & \textit{75.77} \\

-w/o both & 7.94 & 39.05 & -31.11 & 76.51 & 3.17 & 54.92 & -51.75 & 70.95 & 11.75 & 34.60 & -22.86 & 76.83 & \textit{35.24} & \textit{74.76} \\

\bottomrule
\end{tabular}
\caption{Results of the ablation study on the OPS benchmark, with Qwen2-7B-Instruct serving as the base model. 
}
\label{tab:ab}
\end{table*}

\subsection{Comparison of Exploration Behavior}

To illustrate exploration behavior changes caused by perplexity modulation during RL training, we present in Figure~\ref{fig:dynamic} the proportion curves of 'correct' and 'wrong' predictions across low, median, and high perplexity groups. Consistent with previous analyses, vanilla GRPO explores fewer trajectories that identify high-perplexity solutions as correct (as indicated by the green line with the lowest proportion of 'correct' predictions) or low-perplexity solutions as wrong (blue line with the lowest proportion of 'wrong' predictions). In contrast, our perplexity-aware GRPO clearly mitigates this preference, resulting in more balanced exploration behavior, as evidenced by the closer curves for both 'correct' and 'wrong' predictions across perplexity bins.

\section{Related Work}
\subsection{Critiquing Capability of LLMs}
The critiquing capability of LLMs is crucial for enabling scalable annotation of supervised mathematical reasoning demonstrations \cite{wang-etal-2024-math, Relation-Aware, DBLP:conf/acl/ZhangZWZLYLZL25, DBLP:conf/iclr/ZhangHBKKA25, yu-etal-2025-self, DBLP:conf/acl/GaoCXWZLLLZXLC25}, which in turn substantially enhances the multi-step mathematical reasoning performance of LLMs. To assess the critique abilities of current models, several critic-oriented benchmarks have been introduced \cite{DBLP:conf/acl/LinGLLLY24, zheng2025processbenchidentifyingprocesserrors}, consistently revealing that existing LLMs still exhibit unsatisfactory critique performance. To obtain annotated critique data for improving LLMs’ critiquing skills, monte carlo sampling has been employed to efficiently annotate mathematical reasoning processes, estimating correctness based on the success rate of subsequent completions \cite{DBLP:conf/acl/WangLSXDLCWS24, DBLP:conf/acl/ZhangZWZLYLZL25}. Despite these advances, they still focus on binary correctness judgments for reasoning processes, without providing more detailed or nuanced critiques.
To address this limitation and enhance the interpretability of judgments, recent research has increasingly focused on annotating the critiquing process itself \cite{DBLP:conf/acl/GaoCXWZLLLZXLC25, DBLP:conf/iclr/ZhangHBKKA25, yu-etal-2025-self}. For example, \citeauthor{DBLP:conf/acl/GaoCXWZLLLZXLC25} \shortcite{DBLP:conf/acl/GaoCXWZLLLZXLC25} employ GPT-4o to generate step-by-step critique annotations, providing both correctness assessments for each reasoning step and detailed explanations. 
Collectively, these studies aim to create high-quality supervised fine-tuning datasets to train models with improved critiquing capability. However, they pay little attention to delving into the underlying reason for the poor critiquing performance of LLMs.

\subsection{Enhancing LLM Reasoning via RL}

Recent studies have demonstrated that RL is a promising approach for unlocking the long-chain, multi-step reasoning capabilities of LLMs. Such advanced reasoning skills are also essential when LLMs are tasked with performing critiques. To reduce training costs, \citeauthor{DBLP:journals/corr/abs-2402-03300} \shortcite{lu-etal-2024-rethinking} propose Group Relative Policy Optimization (GRPO), which replaces the critic model with a baseline estimated from group scores, thereby enabling large-scale RL. However, \citeauthor{liu2025understanding} \shortcite{liu2025understanding} identify optimization biases inherent in GRPO, such as response-level length bias and question-level difficulty bias. To address these issues, they propose Dr.GRPO, which eliminates response-length normalization and standard deviation normalization, thereby achieving a more unbiased optimization process. Furthermore, several studies \cite{DBLP:journals/corr/Entropy, DBLP:journals/corr/DAPO} highlight the critical role of entropy in RL, which encourages the policy to explore more diverse trajectories. In parallel, \citeauthor{xie2025teaching} \shortcite{DBLP:conf/aaai/TianLZ0CGS025} investigate LLMs' critiquing capability in code generation tasks and demonstrate that training critic models via RL notably enhances the critiquing capability in this domain. These advances show new potential for enhancing the ability of LLMs to critique mathematical reasoning.

\section{Conclusion}
In this paper, we delve into and quantify the underlying reason (i.e., imbalanced evaluation preference) for the poor critiquing performance of LLMs by meticulously constructing a one-to-many problem solution benchmark. To investigate the behavior difference in depth, we conduct a statistical preference analysis oriented on perplexity and find an intriguing phenomenon—LLMs incline to judge solutions with lower perplexity as correct, which is dubbed as the imbalanced evaluation preference. Motivated by the analysis of the potential reason, we propose a novel perplexity-aware reinforcement learning algorithm to rectify the evaluation preference, which supports the LLMs to explore counter-preference trajectories that judge lower perplexity as wrong and higher perplexity as correct. Extensive experimental results on our built OPS and existing available critic benchmarks demonstrate the validity of our method.

\section{Acknowledgments}
This research was funded by the National Natural Science Foundation of China (NO. 62206267).
\bibliography{aaai2026}
\end{document}